\documentclass[conference]{IEEEtran}
\IEEEoverridecommandlockouts

\usepackage{amsmath, amssymb, amsfonts}
\usepackage{algorithm, algorithmic}
\usepackage{graphicx}
\usepackage{textcomp}
\usepackage{xcolor}
\usepackage{tikz}
\usetikzlibrary{shapes.geometric, arrows.meta, positioning, calc, fit}
\usepackage{academicons}
\usepackage[T1]{fontenc}
\usepackage{booktabs}
\usepackage{mathptmx}
\usepackage{comment}
\usepackage{multirow}
\usepackage[hyphens]{url}
\usepackage{hyperref}
\usepackage{float}
\usepackage{microtype}

\newcommand{\orcid}[1]{\href{https://orcid.org/#1}{\textcolor[HTML]{A6CE39}{\aiOrcid}}}

\begin{document}

\def\BibTeX{{\rm B\kern-.05em{\sc i\kern-.025em b}\kern-.08em
    T\kern-.1667em\lower.7ex\hbox{E}\kern-.125emX}}

\title{CW-BASS: Confidence-Weighted Boundary-Aware Learning for Semi-Supervised Semantic Segmentation}

\author{
    \IEEEauthorblockN{Ebenezer Tarubinga}
    \IEEEauthorblockA{
        \textit{Dept. of Artificial Intelligence} \\
        \textit{Korea University} \\
        Seoul, Korea \\
        psychofict@korea.ac.kr
    }
    \and
    \IEEEauthorblockN{Jenifer Kalafatovich}
    \IEEEauthorblockA{
        \textit{Dept. of Artificial Intelligence} \\
        \textit{Korea University} \\
        Seoul, Korea \\
        jenifer@korea.ac.kr
    }
    \and
    \IEEEauthorblockN{Seong-Whan Lee*\thanks{This research was supported by the Institute of Information \& Communica- tions Technology Planning \& Evaluation (IITP) grant, funded by the Korea government (MSIT) (No. RS-2019-II190079 (Artificial Intelligence Graduate School Program (Korea University)), No. RS-2024-00436857 (Information Technology Research Center (ITRC)), and No. RS-2024-00457882 (AI Re- search Hub Project)).
    *Seong-Whan Lee is the corresponding author.}
}
    \IEEEauthorblockA{
        \textit{Dept. of Artificial Intelligence} \\
        \textit{Korea University} \\
        Seoul, Korea \\
        sw.lee@korea.ac.kr
    }
}
\maketitle

\begin{abstract}

Semi-supervised semantic segmentation (SSSS) aims to improve segmentation performance by utilising large amounts of unlabeled data with limited labeled samples. Existing methods often suffer from coupling, where over-reliance on initial labeled data leads to suboptimal learning; confirmation bias, where incorrect predictions reinforce themselves repeatedly; and boundary blur caused by limited boundary-awareness and ambiguous edge cues. To address these issues, we propose CW-BASS, a novel framework for SSSS. In order to mitigate the impact of incorrect predictions, we assign confidence weights to pseudo-labels. Additionally, we leverage boundary-delineation techniques, which, despite being extensively explored in weakly-supervised semantic segmentation (WSSS), remain underutilized in SSSS.
Specifically, our method: (1) reduces coupling via a confidence-weighted loss that adjusts pseudo-label influence based on their predicted confidence scores, (2) mitigates confirmation bias with a dynamic thresholding mechanism that learns to filter out pseudo-labels based on model performance, (3) tackles boundary blur using a boundary-aware module to refine segmentation near object edges, and (4) reduces label noise through a confidence decay strategy that progressively refines pseudo-labels during training.
Extensive experiments on Pascal VOC 2012 and Cityscapes demonstrate that CW-BASS achieves state-of-the-art performance. Notably, CW-BASS achieves a 65.9\% mIoU on Cityscapes under a challenging and underexplored 1/30 (3.3\%) split (100 images), highlighting its effectiveness in limited-label settings. Our code is available at \url{https://github.com/psychofict/CW-BASS}.

\end{abstract}


\section{Introduction}

Semantic segmentation, the task of assigning semantic labels to each pixel in an image, is fundamental to applications like autonomous driving, medical imaging, and scene understanding \cite{shin2024revisiting}. However, it heavily relies on large-scale annotated datasets, which are expensive and labour-intensive to produce \cite{10.1007/978-3-030-59710-8_54}, making it difficult to adapt models to new datasets with limited labeled data \cite{10.1109/cvpr.2019.00262}. To address this, researchers have increasingly explored semi-supervised semantic segmentation (SSSS), which leverages both labeled and unlabeled data through self-training \cite{lee2013pseudo}, consistency-based regularization \cite{laine2017temporal}, and generative models \cite{zhang2022semi} to improve model generalization.

Early semi-supervised methods mainly relied on self-training for its simplicity and effectiveness. Here, a teacher model trained on labeled data generates pseudo-labels for unlabeled data, which are then used to train a student model \cite{10.1007/978-3-030-58601-0_26}. This allows models to learn from large volumes of unlabeled data, improving generalization and robustness, especially in data-scarce domains like medical imaging \cite{liu2022perturbed}. Building on this, more advanced methods such as consistency regularization and generative models \cite{zhang2022semi} have been proposed. Consistency-based approaches enforce stable predictions under input perturbations, while generative models, such as GANs, synthesize training data or enhance pseudo-label quality to boost segmentation performance \cite{cho2024interactive}.

\begin{figure}[t]
    \centering
    \includegraphics[width=\columnwidth, keepaspectratio]{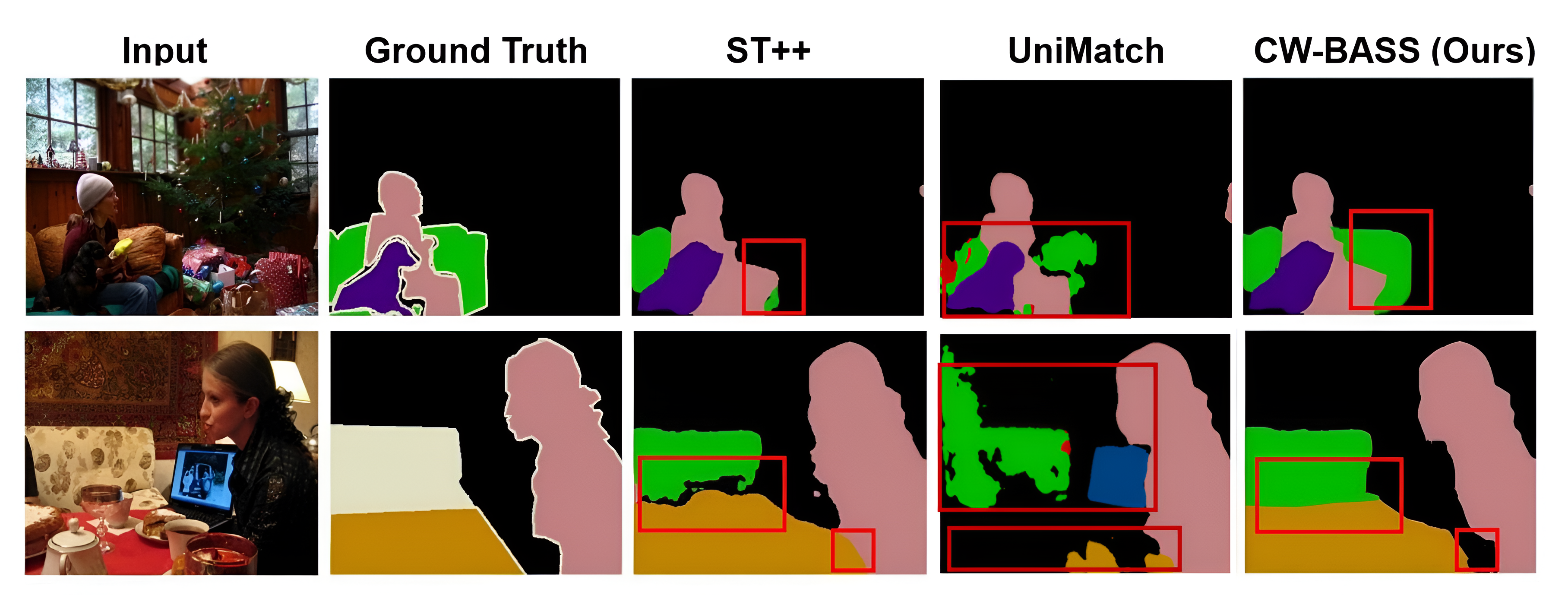}
    \vspace{-1.5em} 
    \caption{Qualitative Comparison with SOTA methods on the Pascal VOC 2012 under 1/8 supervised protocol. Our method, CW-BASS excels in limited label cases as shown above. From left to right: Input Image, Ground Truth, ST++~\cite{yang2022st++}, UniMatch~\cite{yang2023revisiting} and CW-BASS (Ours). Red rectangles highlight regions where our method improves segmentation performance.}
    \label{Fig._1_Mini_Results_Pascal.png}
\end{figure}

Despite these advances, challenges such as coupling, confirmation bias, and boundary delineation persist due to label noise and limited supervision \cite{10.1007/978-3-030-01219-9_18}. Coupling arises when model predictions overly depend on the initial labeled data, limiting pseudo-label quality. Techniques like exponential moving averages \cite{10.1109/tpami.2022.3208419} and consistency regularisation \cite{10.3390/a13010026} have addressed this, but often rely on fixed confidence thresholds that fail to adapt during training. This rigidity can exclude useful low-confidence samples and hinder generalization. Confirmation bias further worsens this by causing models to favour predictions that align with their initial biases, creating a feedback loop where errors in pseudo-labels are repeatedly reinforced. In \cite{zheng2024intramixintraclassmixupgeneration}, it is shown that high intra-class variation leads to poor pseudo-label quality and reduced generalization. 

Boundary blur occurs when a model fails to accurately recognize the correct category of pixels at object boundaries \cite{zheng2024intramixintraclassmixupgeneration}. In weakly-supervised semantic segmentation (WSSS), boundary-delineating methods have been proposed to reduce the effects of inaccurate pixel-level annotations~\cite{chen2022class} whilst their application in SSSS remains relatively under-explored, highlighting a critical gap in the current research landscape. 

To tackle these limitations, we propose a novel dual-stage training framework that integrates dynamic confidence learning and boundary-delineation methods. We specifically solve the aforementioned issues and achieve remarkable segmentation performance as demonstrated in Fig.~\ref{Fig._1_Mini_Results_Pascal.png} 

Our key contributions are summarised as follows:

\begin{enumerate}
    \item \textbf{Confidence-Weighted Loss Function}: We apply a novel confidence-weighted cross-entropy loss function that adjusts pixel contribution to the total loss based on confidence score, enabling more reliable predictions while learning from less certain ones.
    
    \item \textbf{Dynamic Thresholding Mechanism}: We introduce an adaptive thresholding mechanism that updates the pseudo-label confidence threshold according to the model's training performance, ensuring accurate predictions without discarding potentially valuable low-confidence data.

    \item \textbf{Boundary Aware Technique}: We introduce a boundary-aware technique module to enhance the model's ability to learn intricate, fine-grained details, significantly improving segmentation accuracy, especially near object boundaries.
    
    \item \textbf{Confidence Decay Strategy}: We employ a confidence decay strategy that progressively reduces the influence of low-confidence pseudo-labels during training, promoting exploration early on and focusing on high-confidence predictions as the model stabilizes.
\end{enumerate}

The remainder of this paper is organized as follows: Section II reviews related work, Section III describes the proposed method, Section IV presents experiments and results and Section V concludes the paper.

\section{Related Work}

\subsection{Pseudo-Labeling}
Pseudo-labeling is a key component in self-training for semi-supervised learning (SSL), where model predictions on unlabeled data become pseudo-labels for subsequent training~\cite{lee2013pseudo}. Although effective, it is vulnerable to error propagation, where inaccurate pseudo-labels aggravate model biases and reduce performance~\cite{zhang2020semi}. To address this, teacher-student frameworks stabilize pseudo-labeling using an exponential moving average (EMA) of the student’s weights~\cite{tarvainen2017mean}. Methods such as~\cite{yang2022st++} refine the teacher iteratively, while~\cite{shin2024revisiting} reuses earlier teacher predictions to mitigate confirmation bias.

We address these issues by introducing a dynamic thresholding mechanism that adjusts to the model’s evolving confidence during training. Unlike~\cite{yang2022st++} and~\cite{shin2024revisiting}, which use static thresholds, our adaptive threshold improves label quality by filtering low-confidence predictions as the model learns. Fixed thresholds may discard useful uncertain labels or allow noise to propagate. Additionally, our confidence decay strategy gradually reduces the effect of noisy pseudo-labels, enabling initial exploration and later focusing on reliable regions.

\subsection{Consistency Regularization}
Consistency regularization ensures predictions remain stable under input perturbations. Early methods like Temporal Ensembling~\cite{laine2017temporal} and VAT~\cite{miyato2018virtual} promote this by comparing predictions on augmented inputs. Extensions such as T-VAT~\cite{wang2022semi} use adversarial perturbations in a teacher-student setup, while~\cite{cho2024interactive} uses two student networks to refine pseudo-labels through feedback.

These methods often rely on complex augmentations or adversarial training, increasing computational cost. In contrast, we avoid explicit consistency loss. Instead, our confidence-weighting and decay, we inherently dampen the impact of unstable predictions and achieve similar robustness without adversarial training or augmentation-heavy designs.

\subsection{Boundary Refined Semantic Segmentation}
Boundary precision is a known weakness in semantic segmentation, especially under limited supervision. In weakly-supervised semantic segmentation (WSSS), boundary-aware methods such as~\cite{zheng2024intramixintraclassmixupgeneration} \cite{chen2022class} refine coarse Class Activation Maps (CAMs) using spatial priors but often miss fine boundary details due to reliance on image-level labels, limiting their use in SSSS.

In SSSS, boundary-aware methods remain underexplored. The distinction between boundary refinement in WSSS and SSSS is significant. In WSSS, boundaries are inferred from global image-level cues, whereas SSSS requires refining noisy, pixel-level pseudo-labels that already carry structural uncertainty. This makes boundary refinement in SSSS more challenging, and addressing it requires methods like ours that explicitly target spatial inconsistencies in dense predictions. ~\cite{dong2024boundary} learning class prototypes near edges by clustering features, which improves precision but introduces additional complexity and hyperparameter sensitivity.

We propose a simpler, more efficient alternative: Sobel-based boundary detection integrated into the loss function. Sobel filters \cite{jensen2020multiobject} are fast, interpretable, and differentiable enough for training compared to other edge detectors like Canny. They offer a non-learned yet effective method to identify edges, especially valuable when pseudo-label quality is still evolving. We use them to extract edge information and generate binary boundary masks, enabling a boundary-aware loss, which guides the model to focus on spatial inconsistencies, significantly improving segmentation near edges.

\begin{figure*}[ht]
\centering
\includegraphics[width=\textwidth]{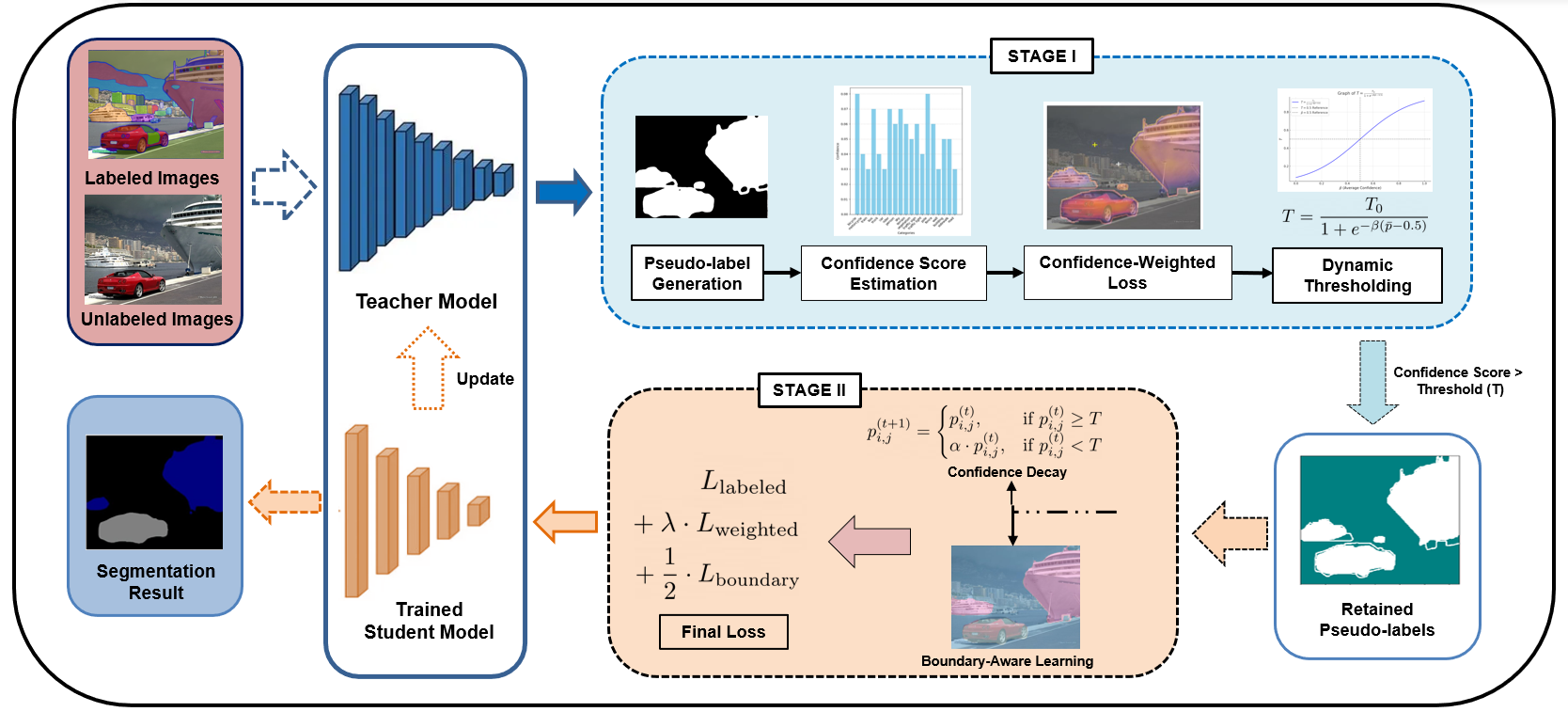}
\caption{Overview of the CW-BASS Framework. In Stage 1, the teacher model generates pseudo-labels with confidence scores for unlabeled data. The confidence-weighted loss and dynamic thresholding filter reliable predictions to train the student model. In Stage 2, a confidence decay strategy and boundary-aware module progressively improve segmentation accuracy near object boundaries.}
\label{Fig._2_Framework.png}
\end{figure*}

\section{Method}


Our proposed method, \textbf{Confidence-Weighted Boundary-Aware Learning (CW-BASS)}, operates in \textbf{two stages}, as shown in Fig.~\ref{Fig._2_Framework.png}. In the \textit{first stage}, the teacher model generates pseudo-labels with confidence scores for unlabeled data, which are used to compute a confidence-weighted loss. A dynamic thresholding mechanism adaptively filters low-confidence pseudo-labels based on training progress. In the \textit{second stage}, a confidence decay strategy reduces the impact of low-confidence pixels, while a boundary-aware module improves accuracy near object edges. The student model is trained using the final loss to produce segmentation results, and the teacher model is updated accordingly.

\subsection{Stage 1: Training the Teacher Model}

\subsubsection{Pseudo-Label Generation with Confidence Estimation}

We address the problem of semantic segmentation in a semi-supervised setting, where a labeled dataset $\mathcal{D}_L = \{ (x_i, y_i) \}_{i=1}^{N_L}$ and an unlabeled dataset $\mathcal{D}_U = \{ x_i \}_{i=1}^{N_U}$ are available, with $N_L$ and $N_U$ representing the number of samples in the labeled and unlabeled datasets, respectively. 

We train a teacher model on $\mathcal{D}_L$, and use it to generate high-quality pseudo-labels for $\mathcal{D}_U$ and train a student model that improves performance on both datasets.

For each unlabeled image $x_i \in \mathcal{D}_U$, the teacher model, ${f}_T$ generates logits, $z_i$:
\begin{equation}
    z_i = f_T(x_i).
\end{equation}

From these logits, we derive \textit{pseudo-labels, $s_i$} and corresponding \textit{pixel-wise confidence scores, $p_i$} from the logits:
\begin{equation}
    s_{i,j} = \arg\max_{k} (z_{i,j,k}),
\end{equation}
\begin{equation}
    p_{i,j} = \max_{k} \left( \text{softmax}(z_{i,j,k}) \right),
\end{equation}
where $s_{i,j}$ is the pseudo-label for pixel $j$ in image $x_i$;  \hfill\\
$p_{i,j} \in [0, 1]$ represents the confidence score for pixel $j$; and \hfill\\
$k$ is the index of the possible classes.

\subsubsection{Confidence-Weighted Loss}
The loss consists of labeled and unlabeled components:

\textbf{Confidence-Weighted Cross-Entropy Loss.} To address the label noise introduced by inaccurate pseudo-labels from the teacher, we propose a confidence-weighted loss that weights the contribution of each pixel to the overall loss based on its confidence score. 
The pixel-wise confidence $p_i$ is raised to a power $\gamma$ to emphasize high-confidence predictions:

The \textbf{loss for an unlabeled image}, $x_i$ is defined as:
\begin{equation}
    \mathcal{L}_{\text{weighted}} = -\frac{1}{N_i} \sum_{j=1}^{N_i} p_{i,j}^\gamma \cdot \log \left( \frac{\exp(z'_{i,j,s_{i,j}})}{\sum_{k} \exp(z'_{i,j,k})} \right),
\end{equation} 
where $N_i$ is the total number of pixels in image $x_i$; \hfill\\
$z'_i = f_S(x_i)$ are the logits predicted by the student model $f_S$; and
$\gamma \geq 0$ controls the degree of emphasis placed on high-confidence predictions. 

\textbf{Loss for Labeled Data.} For labeled data, we use the standard cross-entropy loss:
\begin{equation}
    \mathcal{L}_{\text{labeled}} = -\frac{1}{N_L} \sum_{i=1}^{N_L} \sum_{j=1}^{N_i} \log \left( \frac{\exp(z'_{i,j,y_{i,j}})}{\sum_{k} \exp(z'_{i,j,k})} \right),
\end{equation}
where \textit{$y_{i,j}$} is the ground truth label for pixel \textit{$j$} in image \textit{$x_i$}.

\textbf{Total Loss.} The total loss is a combination of the losses from labeled and unlabeled data:
\begin{equation}
    \mathcal{L}_{\text{total}} = \mathcal{L}_{\text{labeled}} + \lambda \cdot \mathcal{L}_{\text{weighted}},
\end{equation}
where \textit{$\lambda$} is a balancing parameter between the supervised and unsupervised losses.

\subsubsection{Dynamic Thresholding Mechanism} 
We introduce a dynamic threshold mechanism $T$ to adaptively learn and adjust the confidence threshold for filtering low-confidence pseudo-labels, based on the model's performance during training. 

\textbf{Average Confidence Calculation.} For each batch, we calculate the average confidence score $\bar{p}$:
\begin{equation}
    \bar{p} = \frac{1}{N_{\text{batch}}} \sum_{i=1}^{N_{\text{batch}}} \frac{1}{N_i} \sum_{j=1}^{N_i} p_{i,j},
\end{equation}
where $\bar{p}$ is the average confidence.

\textbf{Dynamic Threshold Adjustment.} The base threshold $T$ is then adjusted using a logistic function based on the average confidence. The logistic function ensures smooth threshold adaptation, preventing abrupt changes that could destabilize training. Compared to linear or stepwise adjustments, it balances sensitivity to confidence trends while maintaining stability \cite{atzmon2019adaptive}.

\begin{equation}
T = \frac{T_0}{1 + e^{-\beta(\bar{p} - 0.5)}},
\end{equation}
where {$T_0$} is the initial confidence threshold, $\bar{p}$ is the initial average confidence and \textit{$\beta$} is a hyperparameter controlling the sensitivity of the threshold adjustment.

\textbf{Retaining Pseudo-labels.} We retain pseudo-labels where the confidence score exceeds the threshold:
\begin{equation}
    \text{Retain } s_{i,j} \text{ if } p_{i,j} \geq T.
\end{equation}

\subsection{Stage 2: Training the Student Model}
\subsubsection{Confidence Decay Strategy} To further reduce the influence of persistently low-confidence pixels, we introduce a decay strategy that progressively reduces their impact throughout training:

For low-confidence pixels, we update their confidence scores as:
\begin{equation}
    p_{i,j}^{(t+1)} =
    \begin{cases}
        p_{i,j}^{(t)}, & \text{if } p_{i,j}^{(t)} \geq T \\
        \alpha \cdot p_{i,j}^{(t)}, & \text{if } p_{i,j}^{(t)} < T
    \end{cases},
\end{equation}
where $p_{i,j}^{(t)}$ is the confidence score at epoch $t$ and $\alpha \in (0, 1)$ is the decay factor.

\subsubsection{Boundary-Aware Learning} After the pseudo-labels are refined and filtered through confidence weighting and thresholding, we introduce boundary-aware learning to focus on the more intricate task of boundary delineation and overall improve segmentation performance.

\textbf{Boundary Detection.} We apply Sobel filters to the pseudo-labels $\hat{y}_i$ to detect edges \cite{jensen2020multiobject}. Sobel filters are computationally efficient and do not require additional training, unlike learned edge detectors. While Canny offers finer edges, its non-differentiable nature complicates integration into end-to-end training \cite{rong2014improved}.

The gradient magnitude $E_{n,h,w}$ is computed using Sobel filters, which detect edges by calculating horizontal $G_x$ and vertical $G_y$ intensity gradients. A pixel is considered part of a boundary if its gradient magnitude exceeds zero ($E_{n,h,w}$ > 0) indicating a local intensity change. This threshold simplifies edge detection while maintaining computational efficiency.

Let $G_x$ and $G_y$ be the horizontal and vertical gradients, respectively. The gradient magnitude at pixel, $(h,w)$ in image, $n$ is:
\begin{equation}
    E_{n,h,w} = \sqrt{(G_{x}^{n,h,w})^2 + (G_{y}^{n,h,w})^2}.
\end{equation}

A binary boundary mask is then defined as:
\begin{equation}
\text{BoundaryMask}_{n,h,w} = 
\begin{cases}
    1, & \text{if } E_{n,h,w} > 0 \\
    0, & \text{otherwise},
\end{cases}
\end{equation}

where \( E_{n,h,w} \) is the gradient magnitude at pixel \((h, w)\);
\( G_x^{n,h,w}, G_y^{n,h,w} \) are the pixel gradients at \((h, w)\) derived via Sobel filters and
\( \text{BoundaryMask}_{n,h,w} \) is the binary mask indicating boundaries (\(\)if \(E_{n,h,w} > 0\), else \(0\)).

\textbf{Boundary-Aware Loss.} To emphasize correct predictions at boundaries, we scale the cross-entropy loss by the boundary mask to encourage the model to focus more on object contours:
\begin{equation}
    \mathcal{L}_{\text{boundary}} = \frac{1}{N_i} \sum_{j=1}^{N_i} \text{CE}_{i,j} \cdot \text{BoundaryMask}_{i,j},
\end{equation}
where \textit{$\text{CE}_{i,j}$} is the pixel-wise cross-entropy at pixel $(i,j)$ and $N_i$ is the total number of pixels in image $x_i$.

\subsubsection{Final Combined Loss} The final loss function combines all components:
\begin{equation}
    \mathcal{L}_{\text{final}} = \mathcal{L}_{\text{labeled}} + \lambda \mathcal{L}_{\text{weighted}} + \frac{1}{2} \mathcal{L}_{\text{boundary}}.
\end{equation}

\subsubsection{Model Update} \hfill\\
\textbf{Student Model.} The student model $f_S$ is trained using $\mathcal{D}_U$ and the refined pseudo-labeled data $(x_i, \hat{y}_i)$:
    \begin{equation}
    f_S^{(t)} \gets \arg\min_{f_S} \sum_{(x_i,\hat{y}_i)\in \mathcal{D}_U} \bigl(\mathcal{L}_{\text{weighted}}(f_S(x_i), \hat{y}_i, p_i) + \tfrac{1}{2}\mathcal{L}_{\text{boundary}}\bigr).
    \end{equation}
\textbf{Teacher Model.} Once the student model $f_S^{(t)}$ is trained, we update the teacher model $f_T^{(t)}$ via direct weight copying

    \begin{equation}
    f_T^{(t)} \leftarrow f_S^{(t)}.
    \end{equation}

\begin{figure*}[t]
\centering
\includegraphics[width=\textwidth]{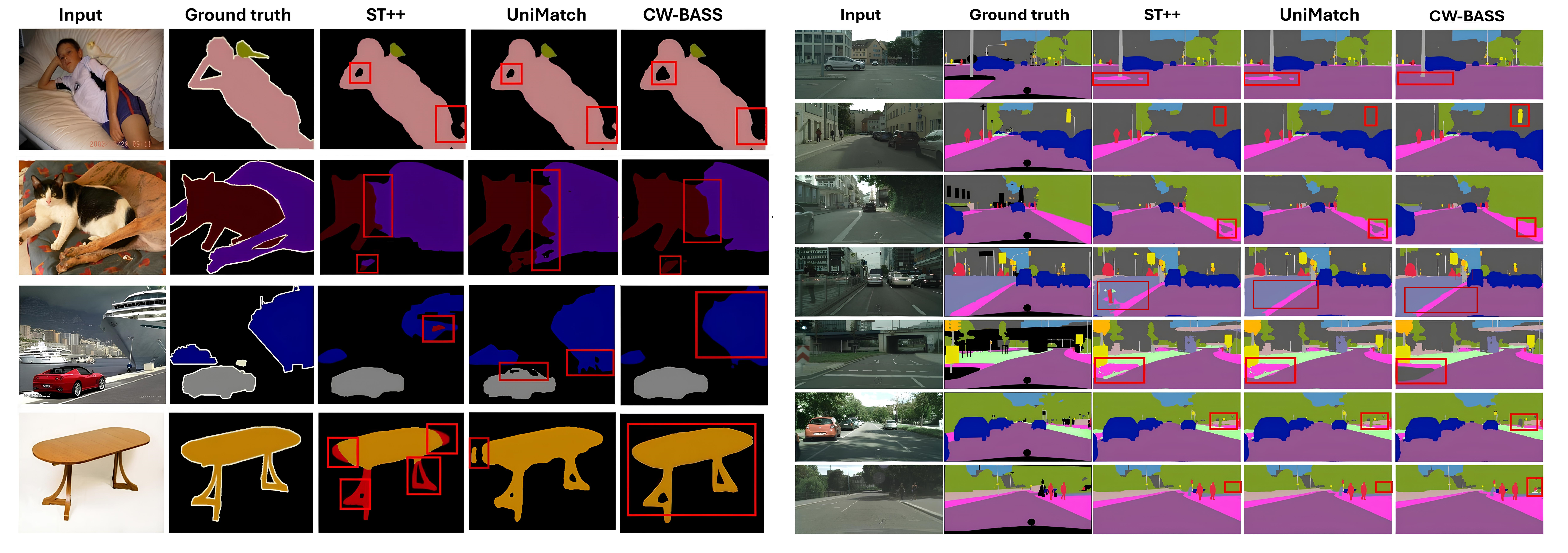}
\caption{Qualitative comparisons of our method, CW-BASS with other state-of-the-art methods, ST++ \cite{yang2022st++} and UniMatch \cite{yang2023revisiting} on the PASCAL VOC 2012 and Cityscapes datasets. Red rectangles highlight regions of improved segmentation performance from the baseline. All methods are compared using ResNet-50 Backbone under the 1/8 setting for validation.}
\label{Fig._4_Qualitative.png}
\end{figure*}

\section{Experiments}

\subsection{Dataset}
For semi-supervised settings, we evaluate our method under various labeled data partitions, such as $1/16$, $1/8$, $1/4$, and $1/2$ of the full labeled dataset.
We perform experiments on two benchmark datasets: Pascal VOC 2012~\cite{Everingham2010} and Cityscapes~\cite{Cordts2016} with comparisons with state-of-the-art methods.

\subsubsection{Pascal VOC 2012} This dataset consists of 1,464 training images and 1,449 validation images, annotated with 21 semantic classes including the background.

\subsubsection{Cityscapes} Cityscapes contains 2,975 training images and 500 validation images, with high-resolution street scenes annotated into 19 classes.

\subsection{Setup}

\subsubsection{Network Architecture} We use the DeepLabV3+ model with ResNet-50 \cite{he2016deep} as the backbone for all experiments, initialized with weights pre-trained on ImageNet \cite{deng2009imagenet}.

\subsubsection{Training Setup}
For optimization, we employ stochastic gradient descent (SGD) with a momentum of 0.9 and a weight decay of \(1 \times 10^{-4}\). The learning rate (LR) and scheduler are as follows: Pascal VOC: 0.001, and for Cityscapes: 0.004 due to larger image sizes and batch adjustments. \hfill\\
A polynomial learning rate decay is used, defined as: \( \text{LR}_{\text{current}} = \text{LR}_{\text{initial}} \times \left(1 - \frac{\text{iter}}{\text{total\_iters}}\right)^{0.9} \). \hfill\\
Batch size for Pascal VOC is set to 16, and for Cityscapes, due to higher resolution, is set to 8. The model is trained for 80 epochs on Pascal and 240 epochs on Cityscapes. To ensure fair comparisons and computational efficiency, we do not employ advanced techniques such as OHEM, auxiliary supervision\cite{chen2021semisupervised}, or SyncBN.

\begin{table*}[t]
\caption{Performance comparison on the Pascal VOC 2012 and Cityscapes datasets in terms of mIoU (\%) across different supervised partitions. \textbf{Best} results are in bold. All experiments used the DeepLabV3+ model with a ResNet-50 backbone. SupOnly* refers to the supervised-only baseline network.}
\label{tab:voc+city_performance}
\begin{center}
\renewcommand{\arraystretch}{1.2} 
\setlength{\tabcolsep}{8pt} 
\begin{tabular}{c | c | c c c c | c c c c}
\hline
\multirow{2}{*}{\textbf{Method}} & \multirow{2}{*}{\textbf{Venue}} & \multicolumn{4}{c|}{\textbf{Pascal VOC 2012}} & \multicolumn{4}{c}{\textbf{Cityscapes}} \\ \cline{3-10} 
 & & 1/16 (92) & 1/8 (183) & 1/4 (366) & 1/2 (732) & 1/30 (100) & 1/16 (186) & 1/8 (372) & 1/4 (744) \\ \hline
SupOnly* & - & 44.00 & 52.30 & 61.70 & 66.70 & 55.10 & 63.30 & 70.20 & 73.12 \\
ST \cite{yang2022st++} & CVPR'22 & 71.60 & 73.30 & 75.00 & – & 60.90 & – & 71.60 & 73.40 \\
ST++ \cite{yang2022st++} & CVPR'22 & 72.60 & 74.40 & 75.40 & – & 61.40 & – & 72.70 & 73.80 \\
\hline
MT \cite{tarvainen2017mean} & NeurIPS'17 & 66.77 & 70.78 & 73.22 & 74.75 & – & 66.14 & 72.03 & 74.47 \\
GCT \cite{ke2020guided} & ECCV'20 & 64.05 & 70.47 & 73.45 & 75.20 & – & 65.81 & 71.33 & 75.30 \\
CPS \cite{chen2021semisupervised} & CVPR'21 & 71.98 & 73.67 & 74.90 & 76.15 & – & 74.47 & 76.61 & 77.83 \\
ELN \cite{kwon2022semi} & CVPR'22 & – & 73.30 & 74.63 & – & – & – & 70.33 & 73.52 \\
PS-MT \cite{liu2022perturbed} & CVPR'22 & 72.80 & 75.70 & 76.02 & 76.64 & – & 74.37 & 76.92 & 77.64 \\
ESL \cite{ma2023enhanced} & ICCV'23 & 61.74 & 69.50 & 72.63 & 74.69 & – & 71.07 & 76.25 & 77.58 \\
UniMatch \cite{yang2023revisiting} & CVPR'23 & 71.90 & 72.48 & 75.96 & \textbf{77.40} & – & 74.03 & 76.77 & 77.49 \\
CPSR \cite{yin2024class} & CVIU'24 & 72.20 & 73.02 & 76.00 & 77.31 & – & 74.49 & \textbf{77.25} & 78.01 \\
DUEB \cite{smita2025uncertainty}  & WACV'25 & 72.41 & 74.89 & 75.85 & 75.94 & - & 72.38 & 76.18 & 77.85 \\
\hline
\textbf{CW-BASS (Ours)} & IJCNN'25 & \textbf{72.80} & \textbf{75.81} & \textbf{76.20} & 77.15 & \textbf{65.87} & \textbf{75.00} & 77.20 & \textbf{78.43} \\
\hline
\end{tabular}
\end{center}
\end{table*}

\begin{table}[ht]
    \centering
    \caption{Comparison on the number of unlabeled data required to train SSSS SOTA methods}
    \label{tab:epochperunlabeled_data}
    \resizebox{\columnwidth}{!}{%
    \begin{tabular}{l|c|cccc|cccc}
        \toprule
        \multirow{2}{*}{Method} & \multirow{2}{*}{Networks} 
            & \multicolumn{4}{c|}{Pascal VOC 2012} 
            & \multicolumn{4}{c}{Cityscapes} \\
        \cline{3-10}
          &   & 1/16 & 1/8 & 1/4 & 1/2 & 1/16 & 1/8 & 1/4 & 1/2 \\
        \midrule
        CPS \cite{chen2021semisupervised} & 2 
            & 10.5k & 10.5k & 10.5k & 10.5k 
            & 2.9k & 2.9k & 2.9k & 2.9k \\
        PS-MT \cite{liu2022perturbed} & 3 
            & 19.8k & 18.5k & 15.8k & 10.5k 
            & 5.5k & 5.2k & 4.4k & 2.9k \\
        ST++ \cite{yang2022st++} & 4 
            & 14.8k & 13.8k & 11.9k & 7.9k 
            & 4.1k & 3.9k & 3.3k & 2.2k \\
        CW-BASS & 2 
            & 9.9k & 9.2k & 7.9k & 5.2k 
            & 2.7k & 2.6k & 2.2k & 1.4k \\
        \bottomrule
    \end{tabular}%
    }
\end{table}

\subsubsection{Implementation Details}
We apply similar semi-supervised training settings as most state-of-the-art-methods to ensure fair comparisons. Data augmentation techniques include random horizontal flipping and random scaling (ranging from 0.5 to 2.0)~\cite{yang2022st++}. We use reduced cropping sizes during training compared to CPS~\cite{chen2021semisupervised}, ie. 321 × 321 pixels on Pascal VOC and 721 × 721 pixels on Cityscapes to save memory. On unlabeled images, we also utilize color jitter with the same intensity as~\cite{Zou2021}, grayscale conversion, Gaussian blur as stated in~\cite{Chen2020}, and Cutout with randomly filled values. All unlabeled images undergo test-time augmentation as well~\cite{chen2021semisupervised}.

\subsubsection{Evaluation Metrics}
We use the standard mean Intersection over Union (mIoU) metric to evaluate segmentation performance. No ensemble techniques were used for all evaluations.

\subsubsection{Hyperparameters}

In our experiments, we use specific hyperparameters with default settings, and their impact is further analyzed in the ablation studies. Confidence Weighting ($\gamma$) is set to a default value of 1.0, to balance the contribution of high- and low-confidence pseudo-labels in the confidence-weighted loss. For Dynamic Thresholding ($T$), the base threshold ($T_0$) is 0.6 (with sensitivity parameter ($\beta$) set to 0.5), allowing the model to learn filter pseudo-labels dynamically based on its performance, thereby mitigating confirmation bias. Lastly, the Confidence Decay Factor ($\alpha$) is set to 0.9, to gradually refine the learning process.

\subsection{Qualitative Analysis}
Fig. \ref{Fig._1_Mini_Results_Pascal.png} and Fig. \ref{Fig._4_Qualitative.png} show visual segmentation outputs produced by our method, CW-BASS on Pascal VOC and Cityscapes dataset, compared against a prominent SOTAs, ST++ \cite{yang2022st++} and UniMatch~\cite{yang2023revisiting}. The qualitative gains are shown, with our method consistently providing clearer object boundaries, more accurate region delineations, and better overall segmentation performance.

\subsection{Quantitative Analysis}

\subsubsection{Comparison on Segmentation Performance} \hfill\\
Our proposed CW-BASS framework achieves state-of-the-art segmentation performance on both Pascal VOC 2012 and Cityscapes, as shown in Table~\ref{tab:voc+city_performance}.

\textbf{Results on Pascal VOC}. Table~\ref{tab:voc+city_performance} compares performance on the Pascal VOC 2012 dataset using various SSSS SOTA methods. Unlike approaches that use multiple teacher networks and various perturbations, our method relies on a single teacher-student model for self-training. To highlight our performance, we also specifically compare our results with ST++, an advanced and simple self-training approach. Using only 1/8 (12.5\%) of the labeled training data, our method produces a mean Intersection over Union (mIoU) of 75.81\%, outperforming existing state-of-the-art approaches. This also exceeds the performance of the fully supervised baseline, \textit{SupOnly} by a remarkable 23.51\%. We also outpeform most SOTA methods in the rest of the splits.

\textbf{Results on Cityscapes}. Our method also outperforms state-of-the-art methods on the Cityscapes dataset as shown in Table~\ref{tab:voc+city_performance}. Cityscapes presents more complex, high-resolution urban environments. Under an extremely limited and rarely used setting, where only 1/30 (3.3\%) of the training data (100 labeled images) is annotated, our method achieves a remarkable mIoU of 65.87\%. This surpasses the supervised baseline by 10.77\%, showcasing our model's efficiency in extremely limited label scenarios and ability to handle complex scene layouts, fine-grained object boundaries, and diverse urban visuals.

\subsubsection{Comparison on Computational Costs and Training} \hfill\\
Figure \ref{Fig._5_Ablation_cityscapes.png} illustrates the convergence behavior and performance of CW-BASS compared to the self-training method ST++ \cite{yang2022st++} on the Cityscapes dataset during the initial 20 epochs, utilizing the 1/16 split for Cityscapes. CW-BASS stabilizes faster due to its confidence-weighted and boundary refinement. In contrast, ST++ requires iterative retraining after pseudo-label selection. Overall, our method combines quick initial convergence with strong performance gains throughout the training process. 

Table~\ref{tab:epochperunlabeled_data} compares data usage and training overhead across methods, focusing on the number of networks, unlabeled data processed per epoch, and total training epochs. CPS uses a dual-network design and treats the labeled set $\mathcal{D}_L$ without ground truth as additional unlabeled data, increasing training volume. PS-MT applies consistency learning with perturbations to both teacher and student networks. CPS trains two networks, processing 10.5k and 2.9k unlabeled images per epoch on Pascal VOC and Cityscapes, respectively. PS-MT uses three networks with strong augmentations, often doubling or tripling unlabeled data (e.g., 19.8k per epoch on 1/16 Pascal) and requires 320–550 epochs on Cityscapes. ST++ trains four networks with slightly less unlabeled data per epoch than PS-MT (14.8k on Pascal, 4.1k on Cityscapes). In contrast, CW-BASS trains two networks with fewer augmentations, using the least unlabeled data (up to 9.9k on Pascal, 2.7k on Cityscapes) and converges in just 80 epochs on Pascal and 240 on Cityscapes, regardless of partition. This balanced design achieves strong performance while avoiding the overhead of heavy augmentations and large ensembles.

\begin{figure}[t]
    \centering
    \includegraphics[height=4cm, keepaspectratio]{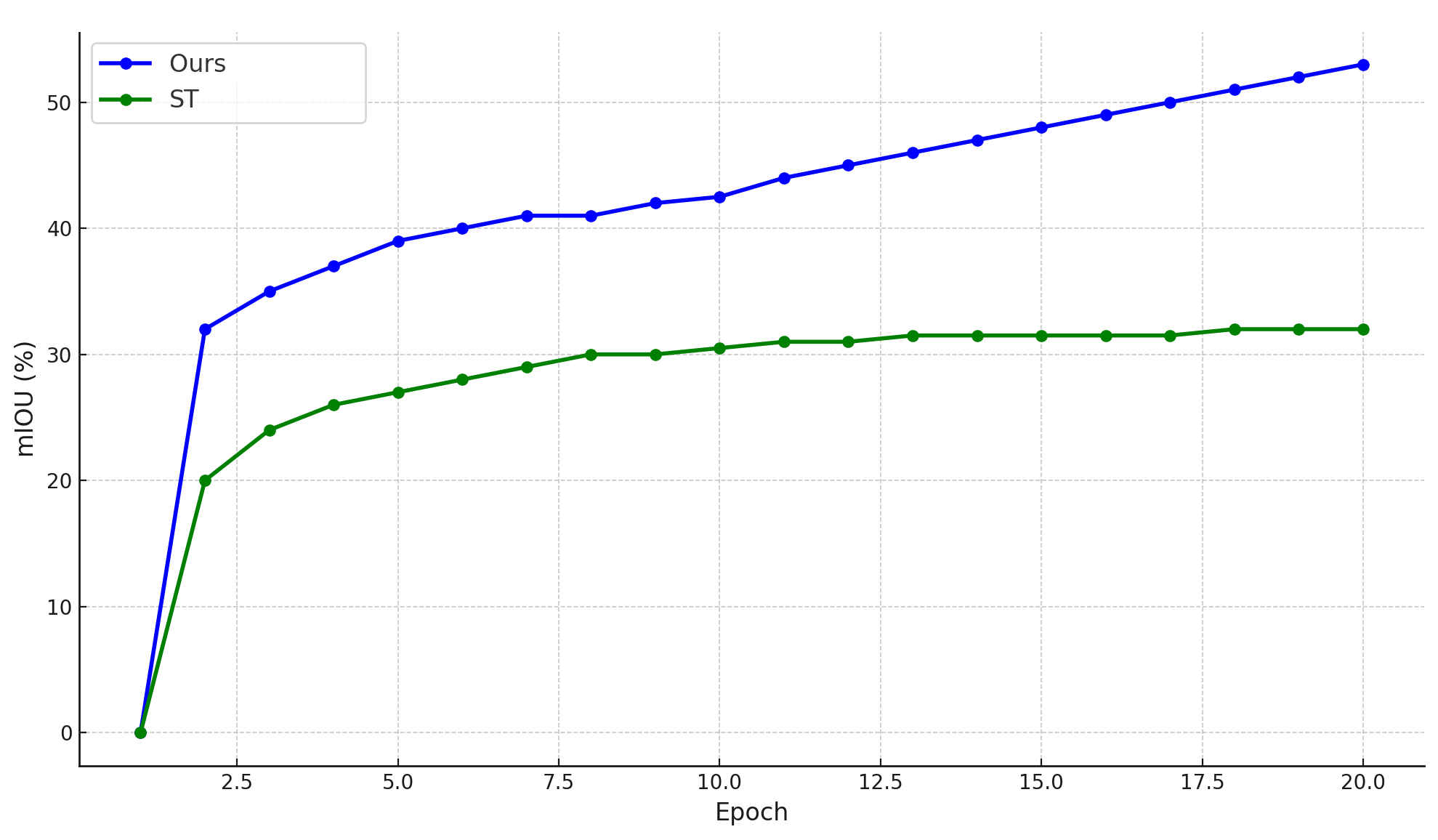}
    \vspace{-0.5em} 
    \caption{Comparison of performance at initial 20 epochs for SOTA methods on Cityscapes 1/16 supervised partition. Our method initializes faster than SOTA methods PS-MT and ST++.}
    \label{Fig._3_Comparison_of_epochs.png}
\end{figure}

\subsection{Ablation Studies}

\begin{figure}[t]
    \centering
    \includegraphics[height=5cm, keepaspectratio]{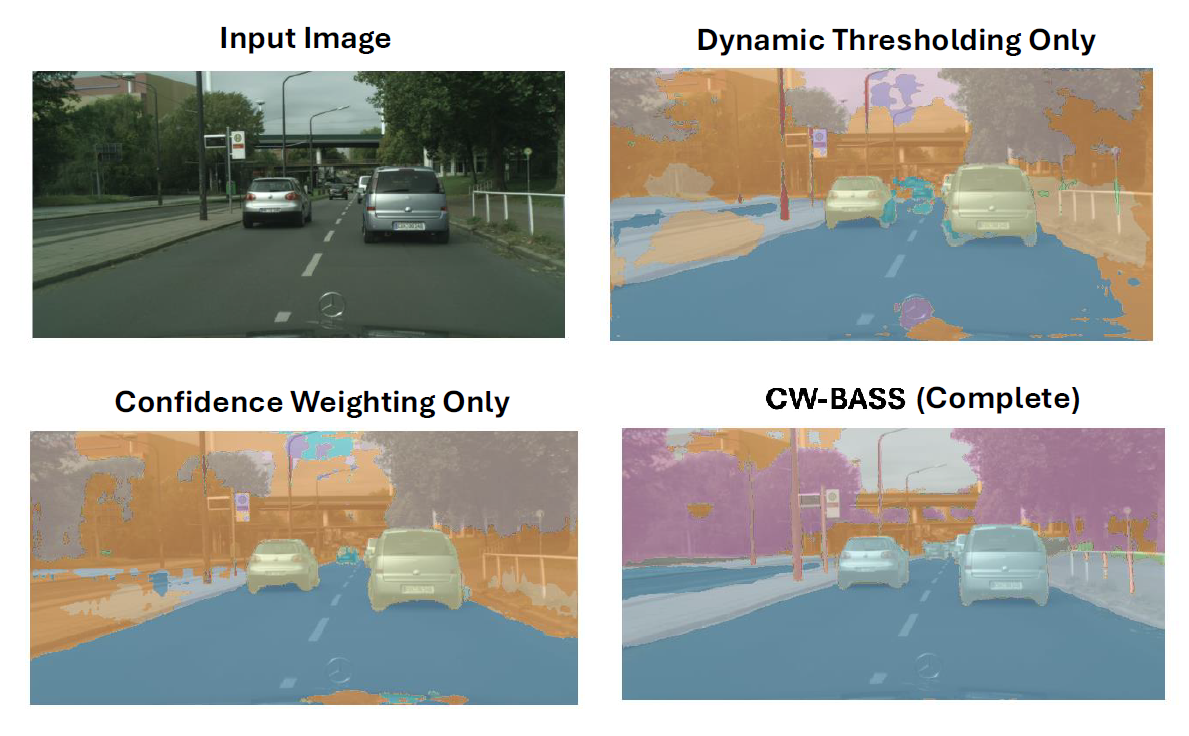}
    \vspace{-0.5em} 
    \caption{Ablation studies comparisons on Cityscapes, 1/8 split. We visualize the predicted mask over the input image.}
    \label{Fig._5_Ablation_cityscapes.png}
\end{figure}

We conduct ablation studies on the various components of our CW-BASS framework to demonstrate their individual effectiveness. Table \ref{tab:ablation-study} shows the effectiveness of each component on the Pascal VOC dataset with a training size of 321 × 321 under a 1/8 (12.5\%) labeled partition. Fig. \ref{Fig._5_Ablation_cityscapes.png} shows the visual comparisons of the predicted masks produced by each component on Cityscapes. We use the standard model (SupOnly) trained with only labeled images as our baseline.

\subsubsection{Ablation Analysis} \hfill\

\textbf{Baseline ($L_{labeled}$).}
We follow standard SSSS practice by using $L_{labeled}$, also referred to as \textit{SupOnly} (Supervised Only), as the baseline for our framework.

\textbf{Confidence-Weighted Loss ($L_{labeled}$ + $L_{weighted}$).}
Confidence-weighting is central to our method, combining $L_{\text{labeled}}$ and $L_{\text{weighted}}$, and achieves an mIoU of 73.43\%. This significantly outperforms the baseline and some SOTA methods, showing how prioritizing high-confidence pseudo-labels mitigates label noise and boosts performance.

\textbf{Boundary-Aware Loss ($L_{boundary}$).}
Adding boundary-aware loss to confidence-weighting improves segmentation near object edges, yielding an mIoU of 74.67\%;a 6.36\% gain over the baseline. This confirms the effectiveness of handling boundary blur with specialized loss functions.

\textbf{Dynamic Thresholding Mechanism ($T$).}
Dynamic thresholding helps filter out low-confidence pseudo-labels. When combined with $L_{\text{labeled}}$ and $L_{\text{weighted}}$, it achieves an mIoU of 69.01\%, demonstrating its role in refining pseudo-label quality during training.

\textbf{Confidence Decay Strategy ($\alpha$).}
This strategy gradually reduces the influence of uncertain pseudo-labels, stabilizing learning over time. The setup ($L_{\text{labeled}} + L_{\text{weighted}} + \alpha$) offers modest gains, showing its value in enhancing pseudo-label robustness.

\textbf{Discussion.}
The study identifies Confidence-Weighted Loss and Boundary-Aware Loss as key contributors to performance. The substantial gain from boundary-aware loss highlights the value of integrating boundary-delineation. Meanwhile, Dynamic Thresholding and Confidence Decay effectively address noisy pseudo-labels and confirmation bias.

\begin{table}[t]
\centering
\caption{Ablation study results over different configurations. The results shown are from the 1/8 supervised partition protocol on Pascal VOC dataset.}
\label{tab:ablation-study}
\begin{tabular}{cccccc}
\toprule
$L_{labeled}$ & $L_{weighted}$ & $\alpha$ & $T$ & $L_{boundary}$ & \text{mIoU(\%)$\uparrow$} \\
\midrule
\checkmark &             &              &            &       & 52.30 \\
\checkmark & \checkmark  & \checkmark   &            &       & 68.31 \\
\checkmark & \checkmark  &              & \checkmark &       & 69.01 \\
\checkmark & \checkmark & \checkmark  &  \checkmark     &                        & 73.43 \\
\checkmark & \checkmark &    &       &     \checkmark        & 74.67 \\
\midrule
\checkmark & \checkmark   & \checkmark  & \checkmark & \checkmark & \textbf{75.81} \\
\bottomrule
\end{tabular}
\end{table}

\begin{table}[t]
\centering
\caption{Ablation study results (MIoU) achieved by standard parameter settings and conservative parameter settings.}
\label{tab:ablation-studyII}
\begin{tabular}{lcc}
\toprule
Hyperparameter group & Hyperparameter values & mIoU (\%)$\uparrow$ \\
\midrule
Standard Settings    & $\gamma=1.0, \beta=0.5, \alpha=0.85$ & 75.81 \\
Conservative Settings & $\gamma=0.5, \beta=1.0, \alpha=1.0$ & 71.40 \\
\bottomrule
\end{tabular}
\end{table}

\subsubsection{Hyperparameter Analysis}

To investigate the impact of our hyperparameter on model performance, we perform experiments using two groups of settings: our standard \& default setting (i.e., $\gamma=1.0, \beta=0.5, \alpha=0.85$), which we use for all our experiments, and a more conservative setting ($\gamma=0.5, \beta=1.0, \alpha=1.0$), as shown in Table \ref{tab:ablation-studyII}. The standard settings batch yield the best model performance because the parameters are balanced, whereas the conservative settings yield a lower mIoU than the standard settings.

For Confidence weighting, $\gamma$ = {0.5, 1.0, 2.0} was tested. A $\gamma$ value of 0.5 reduces the influence of these labels, making the model more conservative to minimize the risk of overfitting to incorrect labels. A $\gamma$ value of 1.0 achieves optimal performance whilst setting $\gamma$ to 2.0 increases the emphasis on high-confidence pseudo-labels resulting in the model to reinforcing confident predictions more aggressively. 

Dynamic thresholding ($T$) involves a base threshold ($T_0$) set at 0.6, which is adjusted using a parameter $\beta$. A higher $\beta$ value (e.g., 1.0) imposes stricter filtering, discarding more low-confidence pseudo-labels. To maintain stability, thresholds are constrained within a range of 0.3 to 0.8, ensuring they do not become excessively lenient or overly strict. 

Finally, the confidence decay factor ($\alpha$), with a default value of 0.9, can be adjusted for specific training needs. Lowering $\beta$ to 0.85 accelerates the decay, which could be particularly useful when a more aggressive reduction of noisy pseudo-labels is required.

\section{Conclusion}
In this work, we introduced Confidence-Weighted Boundary-Aware Semantic Segmentation (CW-BASS), a new SSSS framework that tackles key challenges including confirmation bias, boundary blur, and label noise to achieve state-of-the-art performance. It reduces computational cost by using only two networks, fewer augmentations, and less unlabeled data per epoch, leading to faster convergence with competitive results. This efficient design is well-suited for resource-constrained settings, balancing performance and cost. Additionally, CW-BASS's convergence and strong boundary handling make it practical for real-world applications with limited data and compute.

For future work, our method could explore transformer-based backbones like DINO-ViT to enhance representation learning and pseudo-label reliability in sparse-label scenarios. Broader evaluations on datasets such as COCO and ADE20K would further validate its generalization across diverse, complex scenes.


\end{document}